\documentstyle[aaai]{article}

\begin{document}

\def\pp{\par\noindent}
\newcommand{\ignore}[1]{}

\newcommand{\cmark}{\underline{\hspace*{1em}}}
\newcommand{\tag}[1]{\mbox{\sc #1}}

\newcommand{\ra}{\rightarrow}
\newcommand{\n}{{\cal N}}
\newcommand{\ft}{{\cal F}}
\newcommand{\cl}{{\cal C}}
\newcommand{\hp}{{\cal H}}
\newcommand{\fpr}{{Pr}}
\newcommand{\epr}{{\cal P}}
\newcommand{\pleq}{\prec}
\newcommand{\snow}{{\em SNOW}}
\newcommand{\tbl}{\cal TBL}
\newcommand{\nb}{\cal NB}
\newcommand{\bo}{\cal BO}
\newcommand{\ls}{\cal LS}
\newcommand{\nls}{{\cal S} {\cal N} {\cal O} {\cal W}}

\newcommand{\ppa}{\cal PPA}
\newcommand{\spell}{\cal Spell}
\newcommand{\pos}{\cal POS}
\newcommand{\dl}{\cal p$1$-DL}

\newcommand{\hear}{\mbox{$\{{\it hear}, {\it here}\}$}}
\newcommand{\weather}{\mbox{$\{{\it weather}, {\it whether\/}\}$}}
\newcommand{\desert}{\mbox{$\{{\it desert}, {\it dessert\/}\}$}}
\newcommand{\site}{\mbox{$\{{\it site}, {\it cite\/}, {\it sight\/}\}$}}
\newcommand{\ppset}{\mbox{$\{{\it v}, {\it n\/}\}$}}
\newcommand{\wsd}{\mbox{$\{{\it industrial\/}, {\it living~ organism\/}\}$}}
\newcommand{\their}{\mbox{$\{{\it their}, {\it there}, {\it they're\/}\}$}}

\title{\vspace*{-0.80in}
       {\normalsize \em \hfill Appeared in AAAI-98}\\
       \vspace*{.55in}
Learning to Resolve Natural Language Ambiguities: \\
A Unified Approach}
\author{{\sl Dan Roth} \\
{\normalsize Department of Computer Science}\\
{\normalsize University of Illinois at Urbana-Champaign} \\
{\normalsize Urbana, IL 61801}\\
{\normalsize danr@cs.uiuc.edu}
}

\maketitle
\begin{abstract}

We analyze a few of the commonly used statistics based and machine
learning algorithms for natural language disambiguation tasks and
observe that they can be re-cast as learning linear separators in the
feature space.  Each of the methods makes a priori assumptions, which
it employs, given the data, when searching for its hypothesis.
Nevertheless, as we show, it searches a space that is as rich as the
space of {\em all} linear separators.  We use this to build an
argument for a data driven approach which merely searches for a {\em
good} linear separator in the feature space, without further
assumptions on the domain or a specific problem.

\pp We present such an approach - a sparse network of linear
separators, utilizing the Winnow learning algorithm - and show how to
use it in a variety of ambiguity resolution problems. The learning
approach presented is attribute-efficient and, therefore, appropriate
for domains having very large number of attributes.
\pp

In particular, we present an extensive experimental comparison of our
approach with other methods on several well studied lexical
disambiguation tasks such as context-sensitive spelling correction,
prepositional phrase attachment and part of speech tagging.  In all
cases we show that our approach either outperforms other methods tried
for these tasks or performs comparably to the best.

\end{abstract}

\section{Introduction}

Many important natural language inferences can be viewed as
problems of resolving ambiguity, either semantic or syntactic, 
based on properties of the surrounding context.
\footnotetext{Copyright \copyright\ 1998, American
Association for Artificial Intelligence  (www.aaai.org). All rights
reserved.}
\setcounter{footnote}{0}
Examples include part-of speech tagging, word-sense disambiguation,
accent restoration, word choice selection in machine translation,
context-sensitive spelling correction, word selection in speech
recognition and identifying discourse markers.
In each of these problems it is necessary to disambiguate two or more
[semantically, syntactically or structurally]-distinct forms which have
been fused together 
into the same representation in some medium.  In a prototypical
instance of this problem, word sense disambiguation, distinct semantic
concepts such as {\tt interest rate\/} and {\tt has interest in Math\/}
are conflated in ordinary text. The surrounding context - word
associations and syntactic patterns in this case - are sufficient to
identify the correct form.

Many of these are important stand-alone problems but even more
important is their role in many applications including speech
recognition, machine translation, information extraction and
intelligent human-machine interaction.  Most of the ambiguity
resolution problems are at the lower level of the natural language
inferences chain; a wide range and a large number of ambiguities are
to be resolved simultaneously in performing any higher level natural
language inference.

Developing learning techniques for language disambiguation has been an
active field in recent years and a number of statistics based and
machine learning techniques have been proposed.  A partial list
consists of Bayesian classifiers \cite{gale-word-sense}, decision
lists \cite{Yarowsky94}, Bayesian hybrids \cite{Golding95}, HMMs
\cite{Charniak93}, inductive logic methods \cite{ZelleMo96},
memory-based methods \cite{zdv-97} and transformation-based learning
\cite{Brill95}.  Most of these have been developed in the context of a
specific task although claims have been made as to their applicativity
to others.

In this paper we cast the disambiguation problem as a learning problem
and use tools from computational learning theory to gain some
understanding of the assumptions and restrictions made by different
learning methods in shaping their search space.

The learning theory setting helps in making a few interesting
observations.  We observe that many algorithms, including naive
Bayes, Brill's transformation based method, Decision Lists and the
Back-off estimation method can be re-cast as learning linear
separators in their feature space. As learning techniques for linear
separators these techniques are limited in that, in general, they
cannot learn {\em all} linearly separable functions. Nevertheless, we
find, they still search a space that is as complex, in terms of its VC
dimension, as the space of all linear separators.  This has
implications to the generalization ability of their
hypotheses. Together with the fact that different methods seem to use
different a priori assumptions in guiding their search for the linear
separator, it raises the question of whether there is an alternative -
search for the best linear separator in the feature space, without
resorting to assumptions about the domain or any specific problem.

Partly motivated by these insights, we present a new algorithm, and
show how to use it in a variety of disambiguation tasks.  The
architecture proposed, \snow, is a Sparse Network Of linear separators
which utilizes the Winnow learning algorithm. A target node in the
network corresponds to a candidate in the disambiguation task; all
subnetworks learn autonomously from the same data, in an on line
fashion, and at run time, they compete for assigning the correct
meaning. The architecture is data-driven (in that its nodes are
allocated as part of the learning process and depend on the observed
data) and supports efficient on-line learning. Moreover, The learning
approach presented is attribute-efficient and, therefore, appropriate
for domains having very large number of attributes.  All together, We
believe that this approach has the potential to support, within a
single architecture, a large number of simultaneously occurring and
interacting language related tasks.

To start validating these claims we present experimental results on
three disambiguation tasks.  {\em Prepositional phrase attachment
(\ppa)} is the task of deciding whether the Prepositional Phrase (PP)
attaches to the noun phrase (NP), as in {\tt Buy the car with the
steering wheel} or to the verb phrase (VP), as in {\tt Buy the car
with his money}.  {\em Context-sensitive Spelling correction (\spell)}
is the task of fixing spelling errors that result in valid words, such
as {\tt It's not to late}, where {\tt too\/} was mistakenly typed as
{\tt to}.  {\em Part of speech tagging (\pos)} is the task of
assigning each word in a given sentence the part of speech it assumes
in this sentence.  For example, assign N or V to {\tt talk} in the
following pair of sentences: {\tt Have you listened to his (him) talk
?}.  In all cases we show that our approach either outperforms other
methods tried for these tasks or performs comparably to the best.

This paper focuses on analyzing the learning problem and on motivating
and developing the learning approach; therefore we can only present
the bottom line of the experimental studies and the details are
deferred to companion reports.

\section{The Learning Problem}

Disambiguation tasks can be viewed as general classification
problems. Given an input sentence we would like to assign it a single
property out of a set of potential properties.
Formally, given a sentence $s$ and a predicate $p$ defined on the
sentence, we let $C=\{c_1,c_2,\ldots c_m\}$ be the collection of
possible values this predicate can assume in $s$. It is assumed that
one of the elements in $C$ is the {\em correct} assignment, $c(s,p)$.
$c_i$ can take values from \site~ if the predicate $p$ is the correct
spelling of any occurrence of a word from this set in the sentence; it
can take values from \ppset~ if the predicate $p$ is the attachment of
the PP to the preceding VP ({\tt v}) or the preceding NP ({\tt n}), or
it can take values from \wsd~ if the predicate is the meaning of the
word {\tt plant} in the sentence.  In some cases, such as part of
speech tagging, we may apply a collection $P$ of different predicates
to the same sentence, when tagging the first, second, $k$th word in
the sentence, respectively. Thus, we may perform a classification
operation on the sentence multiple times.  However, in the following
definitions it would suffice to assume that there is a single
pre-defined predicate operating on the sentence $s$; moreover, since
the predicate studied will be clear from the context we omit it and
denote the correct classification simply by $c(s)$.

A {\em classifier} $h$ is a function that maps the set $S$ of all
sentences\footnote{The basic unit studied can be a paragraph or any
other unit, but for simplicity we will always call it a sentence.},
given the task defined by the predicate $p$, to a single value in $C$,
$h:S \ra C$.

In the setting considered here the classifier $h$ is selected by a
training procedure. That is, we assume\footnote{This is usually not
made explicit in statistical learning procedures, but is assumed there
too.}  a class of functions $\hp$, and use the training data to select
a member of this class.  Specifically, given a {\em training corpus}
$S_{tr}$ consisting of {\em labeled example} ($s,c(s)$, 
a learning algorithm selects a hypothesis $h \in \hp$,  the
classifier.

The performance of the classifier is measured
empirically, as the fraction of correct classifications it performs on
a set $S_{ts}$ of test examples. 
Formally,
\begin{equation}
\label{eq:perf}
 Perf(f) = |\{s \in S_{ts} | h(s) = c(s) \}| / |\{s \in S_{ts} \}|.
\end{equation}

A sentence $s$ is represented as a collection of {\em features}, and
various kinds of feature representation can be used.
For example, typical features used in correcting context-sensitive
spelling are {\it context words\/} -- which
test for the presence of a particular
word within $\pm k$ words of the target word, and
{\it collocations} -- which test for a pattern of up to $\ell$ contiguous
words and/or part-of-speech tags around the target word.

It is useful to consider features as sequences of {\em tokens}
(e.g., words in the sentence, or pos tags of the words).  In many
applications (e.g., $n$-gram language models), there is a clear
ordering on the features.  We define here a natural partial order
$\pleq$ as follows: for features $f, g$ define $f \pleq g \equiv f
\subseteq g$, where on the right end side features are viewed simply
as sets of tokens\footnote{There are many ways to define features and
order relations among them (e.g., restricting the number of tokens in a feature,
enforcing sequential order among them, etc.). The following
discussion does not depend on the details; one option is presented 
to make the discussion more concrete.}.  A feature $f$ is of {\em order} $k$ if
it consists of $k$ tokens.

A definition of a disambiguation problem consists of the task
predicate $p$, the set $\cl$ of possible classifications and the set
$\ft$ of features. $\ft^{(k)}$ denotes the features of order $k$.  Let
$|\ft|=n$, and $x_i$ be the $i$th feature. $x_i$ can either be present
({\em active}) in a sentence $s$ (we then say that $x_i=1$), or absent
from it ($x_i=0$).  Given that, a sentence $s$ can be represented as
the set of all {\em active} features in it $s=(x_{i_1},x_{i_2}, \ldots
x_{i_m})$.

From the stand point of the general framework the exact mapping of a
sentence to a feature set will not matter, although it is crucially
important in the specific applications studied later in the paper.  At
this point it is sufficient to notice that the a sentence can be
mapped into a binary feature vector.  Moreover, w.l.o.g we assume that
$|C|=2$; moving to the general case is straight forward.
From now on we will therefore treat classifiers as Boolean functions,
$h:\{0,1\}^n \ra \{0,1\}$.

\section{Approaches to Disambiguation}

Learning approaches are usually categorized as statistical (or
probabilistic) methods and symbolic methods. However, all learning
methods are statistical in the sense that they attempt to make
inductive generalization from observed data and use it to make
inferences with respect to previously unseen data; as such, the
statistical based theories of learning \cite{Vapnik95} apply equally
to both.  The difference may be that symbolic methods do not
explicitly use probabilities in the hypothesis. To stress the
equivalence of the approaches further in the following discussion we
will analyze two ``statistical'' and two ``symbolic''
approaches.

In this section we present four widely used disambiguation methods.
Each method is first presented as known and is then re-cast as a
problem of learning a linear separator.  That is, we show that, there
is a linear condition $\sum_{x_i \in \ft} w_i x_i > \theta$ such that,
given a sentence $s=(x_{i_1},x_{i_2}, \ldots x_{i_m})$, the method
predicts $c=1$ if the condition holds for it, and $c=0$ otherwise.

Given an example $s=(x_1,x_2 \ldots x_m)$ a probabilistic classifier
$h$ works by choosing the element of $C$ that is most probable, that
is $h(s) = argmax_{c_i \in C} Pr(c_i | x_1,x_2,\ldots x_m)\footnote{As
usual, we use the notation $Pr(c_i | x_1,x_2,\ldots x_m)$ as a
shortcut for $Pr(c=c_i|x_1=a_1,x_2=a_2,\ldots x_m=a_m).$},$ where the
probability is the empirical probability estimated from the labeled
training data.  In general, it is unlikely that one can estimate the
probability of the event of interest $(c_i | x_1,x_2,\ldots x_m)$
directly from the training data. There is a need to make some
probabilistic assumptions in order to evaluate the probability of this
event indirectly, as a function of ``more frequent'' events whose
probabilities can be estimated more robustly. Different probabilistic
assumptions give rise to difference learning methods and we describe
two popular methods below.

\subsubsection{The naive Bayes estimation (\nb)}
The naive Bayes estimation (e.g., \cite{DudaHa73}) assumes that given
the class value $c \in C$ the features values are statistically
independent. With this assumption and using Bayes rule the Bayes optimal
prediction is given by:
$ h(s) = argmax_{c_i \in C} \Pi_{i=1}^{m} Pr(x_j | c_i) P(c_i).$

The prior probabilities $p(c_i)$ (i.e., the fraction of training
examples labeled with $c_i$) and the conditional probabilities $Pr(x_j
| c_i)$ (the fraction of the training examples labeled $c_i$ in which
the $j$th feature has value $x_j$) can be estimated from the training
data fairly robustly\footnote{Problems of sparse data may arise,
though, when a specific value of $x_i$ observed in testing has
occurred infrequently in the training, in conjunction with
$c_j$. Various {\em smoothing} techniques can be employed to get more
robust estimations but these considerations will not affect our
discussion and we disregard them.}, giving rise to the naive Bayes
predictor. According to it, the optimal decision is $c=1$ when
$$P(c=1) \Pi_{i} P(x_i|c=1) / P(c=0) \Pi_{i} P(x_i|c=0) > 1.$$
Denoting  $p_i \equiv P(x_i=1|c=1), q_i \equiv P(x_i=1|c=0)$, 
$P(c=r) \equiv P(r)$,
we can write this condition as
$$\frac{P(1) \Pi_{i} p_i^{x_i} (1-p_i)^{1-x_i}}
       {P(0) \Pi_{i} q_i^{x_i} (1-q_i)^{1-x_i}} = 
  \frac{P(1) \Pi_{i} (1-p_i)(\frac{p_i}{1-p_i})^{x_i}}
       {P(0) \Pi_{i} (1-q_i)(\frac{q_i}{1-q_i})^{x_i}} > 1,$$
and by taking log we get that using naive Bayes estimation we predict 
$c=1$ if and only if

$$\log{\frac{P(1)}{P(0)}}  + \sum_{i} \log{ \frac{1-p_i}{1-q_i} } +
  \sum_{i}(\log{\frac{p_i}{1-p_i}} \frac{1-q_i}{q_i}) x_i > 0.$$
We conclude that the decision surface of the naive Bayes algorithm is given by a 
linear function in the feature space. Points which reside on one side of the 
hyper-plane are more likely to be labeled $1$  and points on the other 
side are more likely to be labeled $0$. 

This representation immediately implies that this predictor is optimal
also in situations in which the conditional independence assumption
does no hold.  However, a more important consequence to our discussion
here is the fact that not all linearly separable functions can be
represented using this predictor \cite{Roth98}.

\subsubsection{The back-off estimation (\bo)}
Back-off estimation is another method for estimating the conditional
probabilities $Pr(c_i|s)$. It has been used in many disambiguation
tasks and in learning models for speech recognition
\cite{Katz87,ChenGo96,CollinsBr95}.  The back-off method suggests to
estimate $Pr(c_i|x_1,x_2,\ldots,x_m)$ by interpolating the more robust
estimates that can be attained for the conditional probabilities of
more general events. Many variation of the method exist;
we describe a fairly general one and then present the version used in
\cite{CollinsBr95}, which we compare with experimentally.

When applied to a disambiguation task, \bo\ assumes that the sentence
itself (the basic unit processed) is a feature\footnote{The
assumption that the maximal order feature is the classified sentence
is made, for example, in \cite{CollinsBr95}. In general, the method
deals with multiple features of the maximal order by assuming their
conditional independence, and superimposing the \nb\ approach.} of maximal
order $f = f^{(k)} \in\ft$. We estimate
$$Pr(c_i|s) = Pr(c_i|f^{(k)}) = 
  \sum_{\{f \in \ft | f \pleq f^{(k)} \}} \lambda_{f} Pr(c_i | f).$$
The sum is over all features $f$ which are more general (and thus
occur more frequently) than $f^{(k)}$. The conditional probabilities
on the right are empirical estimates measured on the training
data, and the coefficients $\lambda_f$ are also estimated given the
training data. (Usually, these are maximum likelihood estimates
evaluated using iterative methods, e.g. \cite{Samuelsson96}).

Thus, given an example $s=(x_1,x_2 \ldots x_m)$
the \bo\ method  predicts $c=1$ if and only if
$$ \sum_{i=1}^{|\ft|} \lambda_{i} (Pr(c=1 | x_i) - Pr(c=0 | x_i)) x_i
> 0,$$ a linear function over the feature space.

For computational reasons, various simplifying assumptions are made in
order to estimate the coefficients $\lambda_f$; we describe here the
method used in \cite{CollinsBr95}\footnote{There, the empirical ratios are
smoothed; experimentally, however, this yield only a slight
improvement, going from 83.7\% to 84.1\% so we present it here in the
pure form.}.
We denote by $\n(f^{(j)})$ the number of occurrences
of the $j$th order feature $f^{(j)}$ in the training data. Then \bo\
estimates $P= Pr(c_i |f^{(k)})$ as follows:

\begin{figure}[h]
{\small
If $\n(f^{(k)}) > 0$,  
\hspace*{0.75in} $P  = \fpr(c_i |f^{(k)})$ \\
Else if $\sum_{f \in \ft^{(k-1)}} \n(f) > 0$,
$P = \frac{1}{|\ft^{(k-1)}|} \sum \fpr(c_i |f^{(k-1)})$ \\
Else if $\ldots$ \\
Else if $\sum_{f \in \ft^{(1)}} \n(f) > 0$, 
\hspace*{0.1in} $P  = \frac{1}{|\ft^{(1)}|} \sum \fpr(c_i |f^{(1)})$
}
\end{figure}
In this case, it is easy to write down the linear separator defining
the estimate in an explicit way.  Notice that with this estimation,
given a sentence $s$, only the highest order features active in it are
considered.  Therefore, one can define the weights of the
$j$th order feature in an inductive way, making sure that it is larger
than the sum of the weights of the smaller order features.
Leaving out details, it is clear that we get 
a simple representation of a linear separator over the
feature space, that coincides with the \bo\ algorithm.

It is important to notice that the assumptions made in the \bo\ estimation
method result in a linear decision surface that is, in general, {\em different}
from the one derived in the \nb\ method.

\subsubsection{Transformation Based Learning (\tbl)}
Transformation based learning \cite{Brill95} is a machine learning
approach for rule learning. It has been applied to a number of natural
language disambiguation tasks, often achieving state-of-the-art
accuracy.

The learning procedure is a mistake-driven algorithm that produces a
set of rules. Irrespective of the learning procedure used to derive
the \tbl\ representation,  we focus here on the
final hypothesis used by \tbl\ and how it is evaluated, given an
input sentence, to produce a prediction. We assume, w.l.o.g, 
$|C|=2$.

The hypothesis of \tbl\ is an ordered list of transformations.  A {\em
transformation} is a rule with an antecedent $t$ and a
consequent\footnote{The consequent is sometimes described as a {\em
transformation} $c_i \ra c_j$, with the semantics -- if the current
label is $c_i$, relabel it $c_j$. When $|C|=2$ it is equivalent to
simply using $c_j$ as the consequent.}  $c \in C$.  The antecedent $t$
is a condition on the input sentence.  For example, in \spell, a
condition might be {\tt word W occurs within $\pm k$ of the target
word}. That is, applying the condition to a sentence $s$ defines a
feature $t(s) \in \ft$.  Phrased differently, the application of the
condition to a given sentence $s$, checks whether the corresponding
feature is active in this sentence.  The condition holds if and only
if the feature is active in the sentence.

An ordered list of transformations (the \tbl\ hypothesis), 
is evaluated as follows: 
given a sentence $s$, an initial label $c \in C$ is assigned
to it. Then, each rule is applied, in order, to the
sentence. If the feature defined by the condition of the rule applies,
the current label is replaced by the label in the consequent.  This
process goes on until the last rule in the list is evaluated.  The
last label is the output of the hypothesis.

In its most general setting, the \tbl\ hypothesis is
not a classifier \cite{Brill95}. The reason is that the
truth value of the condition of the $i$th rule may change while
evaluating one of the preceding rules. However, in many 
applications and, in particular, in \spell\ \cite{ManguBr97} and
\ppa\ \cite{BrillRe94} which we discuss later, this is not the
case. There, the conditions do not depend on the labels, and therefore
the output hypothesis of the \tbl\ method can be viewed  as a
classifier. The following analysis applies only for this
case.

Using the terminology introduced above, let
$(x_{i_1},c_{i_1}),(x_{i_2},c_{i_2}), \ldots (x_{i_k},c_{i_k})$ be
the ordered sequence of rules defining the output hypothesis of \tbl.
(Notice that it is quite possible, and happens often
in practice, for a feature to appear more than once in this
sequence, even with different consequents).
While the above description calls for evaluating the hypothesis by
sequentially evaluating the conditions, it is easy to see that 
the following simpler procedure is sufficient:
\begin{center}
Search the ordered sequence in a reversed order. Let $x_{i_j}$ be the
first active feature in the list (i.e., the largest $j$).  Then the
hypothesis predicts $c_{i_j}$.
\end{center}
Alternatively, the \tbl\ hypothesis can be represented as a (positive)
$1$-Decision-List (\dl) \cite{Rivest87}, over the set $\ft$ of
features\footnote{Notice, the order of the features is reversed. Also,
multiple occurrences of features can be discarded, leaving only the
last rule in which this feature occurs. By ``positive'' we mean that
we never condition on the absence of a feature, only on its
presence.}.
\begin{figure}[h]
{\small
\begin{minipage}{\textwidth}
If \hspace*{0.40in} $x_{i_k}$ is active then predict $c_k$. \\
Else 
\hspace*{0.25in} If $x_{i_{k-1}}$ is active then predict $c_{k-1}$. \\
Else       $\ldots$ \\
Else
\hspace*{0.25in} If $x_{1}$ is active then predict $c_1$. \\
Else 
\hspace*{0.25in} Predict the initial value
\end{minipage}
\caption{\tbl~ as a p$1$-Decision List}
\label{fig:tbl}
}
\end{figure}
Given the \dl\ representation (Fig~\ref{fig:tbl}), we can now
represent the hypothesis as a linear separator over the set $\ft$ of
features.  For simplicity, we now name the class labels $\{-1,+1\}$
rather than $\{0,1\}$. Then, the hypothesis predicts $c=1$ if and only
if $\sum_{j=1}^k 2^{j} \cdot c_{i_j} \cdot x_{i_j} >0.$ Clearly, with
this representation the active feature with the highest index
dominates the prediction, and the representations are
equivalent\footnote{In practice, there is no need to use this
representation, given the efficient way suggested above to evaluate
the classifier. In addition, very few of the features in $\ft$ are
active in every example, yielding more efficient evaluation
techniques (e.g., \cite{Valiant98})}.

\subsubsection{Decision Lists (\dl)}
It is easy to see (details omitted), that the above
analysis applies to \dl, a method used, for example, in
\cite{Yarowsky95}. The \bo\ and \dl\ differ only in that they keep
the rules in reversed order, due to different evaluation methods.

\section{The Linear Separator Representation}
\label{sec:ls}

To summarize, we have shown:

\pp
{\bf claim:} {\em 
All the methods discussed -- \nb, \bo, \tbl~ and \dl~ search for a
decision surface which is a linear function in the feature space.
}

This is not to say that these methods assume that the data is linearly
separable.  Rather, all the methods assume that the feature space is
divided by a linear condition (i.e., a function of the form $\sum_{x_i
\in \ft} w_i x_i > \theta$) into two regions, with the property that
in one of the defined regions the more likely prediction is $0$ and in
the other, the more likely prediction is $1$.

As pointed out, it is also instructive to see that these methods yield
{\em different} decision surfaces and that they cannot represent every linearly
separable function.
\section{Theoretical Support for the Linear Separator Framework}
\label{sec:theory}

In this section we discuss the implications these observations have from
the learning theory point of view.

In order to do that we need to resort to some of the basic ideas that
justify inductive learning.  Why do we hope that a classifier learned
from the training corpus will perform well (on the test data) ?
Informally, the basic theorem of learning theory
\cite{Valiant84,Vapnik95} guarantees that, if the training data and
the test data are sampled from the same distribution\footnote{This is
hard to define in the context of natural language; typically, this is
understood as texts of similar nature; see a discussion of this issue
in \cite{GoldingRo96}.}, good performance on the training corpus
guarantees good performance on the test corpus.

If one knows something about the model that generates the data, then
estimating this model may yield good performance on future
examples. However, in the problems considered here, no reasonable
model is known, or is likely to exist. (The fact that the assumptions
discussed above disagree with each other, in general, may be viewed as
a support for this claim.)

In the absence of this knowledge a learning method merely attempts to
make correct predictions. Under these conditions, it can be shown that
the error of a classifier selected from class ${\cal H}$ on
(previously unseen) test data, is bounded by the sum of its training
error and a function that depends linearly on the complexity of ${\cal
H}$.  This complexity is measured in terms of a combinatorial
parameter - the VC-dimension of the class ${\cal H}$ \cite{Vapnik82} -
which measures the richness of the function class.  (See
\cite{Vapnik95,KearnsVa94}) for details).
 
We have shown that all the methods considered here look for a linear
decision surface. However, they do make further assumptions which seem
to restrict the function space they search in.  To quantify this line
of argument we ask whether the assumptions made by the different
algorithms significantly reduce the complexity of the hypothesis
space.  The following claims show that this is not the case; the VC
dimension of the function classes considered by all methods are as
large as that of the full class of linear separators.
\pp 
{\bf Fact $1$:} {\em The VC dimension of the class of linear
separators over $n$ variables is $n+1$.}
\pp
{\bf Fact $2$:} {\em The VC dimension of the class of \dl~ over $n$
variables\footnote{In practice, when using \dl~ as the hypothesis
class (i.e., in \tbl) an effort is made to discard many of the
features and by that reduce the complexity of the space; however, this
process, which is data driven and does not a-priori restrict the
function class can be employed by other methods as well (e.g.,
\cite{Blum95}) and is therefore orthogonal to these arguments.} is
$n+1$.}
\pp 
{\bf Fact $3$:} {\em The VC dimension of the class of linear
separators derived by either \nb\ or \bo\ over $n$ variables is
bounded below by $n$.}

Fact $1$ is well known; $2$ and $3$ can be derived directly from the
definition \cite{Roth98}.

The implication is that a method that merely searches for the optimal
linear decision surface given the training data may, in general,
outperform all these methods also on the test data.  This argument can
be made formal by appealing to a result of \cite{KearnsSc94}, which
shows that even when there is no perfect classifier, the optimal
linear separator on a polynomial size set of training examples is
optimal (in a precise sense) also on the test data.

The optimality criterion we seek is described in Eq. $1$. A linear
classifier that minimizes the number of disagreements (the sum of the
false positives and false negatives classifications).  This task,
however, is known to be NP-hard \cite{HoffgenSi92}, so we need to
resort to heuristics. In searching for good heuristics we are guided
by computational issues that are relevant to the natural language domain.
An essential property of an algorithm is being feature-efficient.
Consequently, the approach describe in the next section makes use of
the {\em Winnow\/} algorithm which is known to produce good results
when a linear separator exists, as well as under certain more relaxed
assumptions \cite{Littlestone91}.
\section{The  \snow\ Approach}
The \snow\ architecture is a network of threshold gates.  Nodes in the
first layer of the network are allocated to input features in a
data-driven way, given the input sentences. Target nodes (i.e., the
element $c \in C$) are represented by nodes in the second layer. Links
from the first to the second layer have weights; each target node is
thus defined as a (linear) function of the lower level nodes.  (A
similar architecture which consists of an additional layer is
described in \cite{GoldingRo96}. Here we do not use the ``cloud''
level described there.)

For example, in \spell, target nodes represent members of the
confusion sets; in \pos, target nodes correspond to different pos
tags.  Each target node can be thought of as an autonomous network,
although they all feed from the same input. The network is {\em
sparse} in that a target node need not be connected to all nodes in
the input layer. For example, it is not connected to input nodes
(features) that were never active with it in the same sentence, or it
may decide, during training to disconnect itself from some of the
irrelevant inputs.

Learning in \snow\ proceeds in an on-line fashion\footnote{Although
for the purpose of the experimental study we do not update the network
while testing.}.  Every example is treated autonomously by
each target subnetworks.
Every labeled example is
treated as positive for the target node corresponding to its label,
and as negative to all others.  Thus, every example is used once by
all the nodes to refine their definition in terms of the others
and is then discarded.  At prediction time, given an input sentence
which activates a subset of the input nodes, the information
propagates through all the subnetworks; the one which produces the
highest activity gets to determine the prediction.

A local learning algorithm, Winnow \cite{Littlestone88}, is used at
each target node to learn its dependence on other nodes.  Winnow is a
mistake driven on-line algorithm, which updates its weights in a
multiplicative fashion.  Its key feature is that the number of
examples it requires to learn the target function grows linearly with
the number of {\em relevant\/} attributes and only logarithmically
with the total number of attributes.  Winnow was shown to learn
efficiently any linear threshold function and to be robust in the
presence of various kinds of noise, and in cases where no
linear-threshold function can make perfect classifications and still
maintain its abovementioned dependence on the number of total and
relevant attributes \cite{Littlestone91,KivinenWa95}.

Notice that even when there are only two target nodes and the cloud
size \cite{GoldingRo96} is $1$ \snow\ behaves differently than pure
Winnow. While each of the target nodes is learned using a positive
Winnow algorithm, a winner-take-all policy is used to determine the
prediction.  Thus, we do not use the learning algorithm here simply as
a discriminator.  One reason is that the \snow\ architecture,
influenced by the Neuroidal system \cite{Valiant94}, is being used in
a system developed for the purpose of learning knowledge
representations for natural language understanding tasks, and is being
evaluated on a variety of tasks for which the node allocation process
is of importance.

\section{Experimental Evidence}
In this section we present experimental results for three of the most
well studied disambiguation problems, \spell, \ppa\ and \pos.  
We present here only the bottom-line results of an extensive
study that appears in companion reports 
\cite{GoldingRo98,KrymolovskyRo98,RothZe98}.

\subsubsection{Context Sensitive Spelling Correction}
Context-sensitive spelling correction is the task of fixing spelling
errors that result in valid words, such as {\it It's not to late},
where {\it too\/} was mistakenly typed as {\it to}.

We model the ambiguity among words by {\it confusion
sets}.  A confusion set $C = \{c_1, \ldots, c_n\}$ means that each
word $c_i$ in the set is ambiguous with each other word. 
All the results reported here use the same pre-defined set of
confusion sets \cite{GoldingRo96}.

We compare \snow\ against \tbl\ \cite{ManguBr97} and a
naive-Bayes based system (\nb).  The latter system
presents a few augmentations over the simple naive Bayes (but still
shares the same basic assumptions) and is among the most successful
methods tried for the problem \cite{Golding95}.
An indication that a Winnow-based algorithm performs well on this
problem was presented in \cite{GoldingRo96}. However, the system
presented there was more involved than \snow~ and allows
more expressive output representation than we allow here.
The output representation of all the approaches compared is
a linear separator.

The results presented in Table~\ref{tab:spell} for \nb\ and \snow\ are
the (weighted) average results of $21$ confusion sets, 19 of them are
of size $2$, and two 
of size $3$. The results presented for the \tbl\footnote{Systems are
compared on the same feature set. \tbl\ was also used with an enhanced
feature set \cite{ManguBr97} with improved results of 93.3\% but we
have not run the other systems with this set of features.} method are
taken from \cite{ManguBr97} and represent an average on a subset of
$14$ of these, all of size $2$.

\begin{table}
\caption{{\bf \spell~ System comparison}.
The second column gives the number of test cases.
All algorithms were trained on 80\% of Brown and tested on the other 20\%;
Baseline simply identifies the most common member of the confusion set
during training, and guesses it every time during testing.
}
\begin{center}
\begin{tabular}{||l c c c c c ||}
\hline
Sets            & Cases & Baseline & \nb  & \tbl & \snow  \\

\hline
{\bf 14} & 1503  & 71.1     & 89.9 & 88.5 & 93.5  \\
{\bf 21} & 4336  & 74.8     & 93.8 &      & 96.4   \\
\hline
\end{tabular}
\vspace{-0.3in}
\end{center}
\label{tab:spell}
\end{table}

\subsubsection{Prepositional Phrase Attachment}
The problem is to decide whether the Prepositional Phrase (PP) attaches 
to the noun phrase, as in  {\tt Buy the car with the steering wheel} 
or the verb phrase, as in {\tt Buy the car with his money}.
Earlier works on this problem \cite{RRR94,BrillRe94,CollinsBr95}
consider as input the four head
words involved in the attachment - the VP head, the first NP head, the
preposition and the second NP head (in this case, {\tt buy, car, with}
and {\tt steering wheel}, respectively).  These four-tuples, along
with the attachment decision constitute the labeled input
sentence and are used to 
generate the feature set. The features recorded are all sub-sequences
of the $4$-tuple, total of $15$ for every input sentence.
The data set used by all the systems in this in this comparison was
extracted from the Penn Treebank WSJ corpus by \cite{RRR94}. It
consists of 20801 training examples and 3097 separate test
examples. In a companion paper we describe an extensive set of
experiments with this and other data sets, under various conditions.
Here we present only the bottom line results that provide direct
comparison with those available in the literature\footnote{\snow\ 
was evaluated with an enhanced
feature set \cite{KrymolovskyRo98} with improved results of 84.8\%.
\cite{CollinsBr95} reports results of 84.4\% on a different enhanced 
set of features, but other  systems were not evaluated on these sets.}.
The results presented in Table~\ref{tab:pp} for \nb\ and \snow\ are
the results of our system on the 3097 test examples. 
The results presented for the \tbl\ and \bo\ are on the same data set,
taken from \cite{CollinsBr95}.

\begin{table}
\caption{{\bf \ppa~ System comparison}.  
All algorithms were trained on 20801 training examples from the WSJ
corpus tested 3097 previously unseen examples from this corpus; all
the system use the same feature set.}
\begin{center}
\begin{tabular}{||l c c c c c ||}
\hline
Test  & Baseline & \nb  & \tbl & \bo & \snow  \\
cases &          &      &      &     &       \\
\hline
3097  & 59.0     & 83.0 & 81.9 & 84.1 & 83.9  \\
\hline
\end{tabular}
\vspace{-0.3in}
\end{center}
\label{tab:pp}
\end{table}

\subsubsection{Part of Speech Tagging}

A part of speech tagger assigns each word in a sentence the part of
speech that it assumes in that sentence. 
See \cite{Brill95} for a survey of much of the work that has been done
on \pos~ in the past few years.
Typically, in English there will be
between $30$ and $150$ different parts of speech depending on the
tagging scheme. In the study presented here, following \cite{Brill95}
and many other studies there are $47$ different tags.  Part-of-speech
tagging suggests a special challenge to our approach, as the problem
is a multi-class prediction problem \cite{RothZe98}.  
In the \snow\ architecture, we
devote one linear separator to each pos tag and each sub network
learns to separate its corresponding pos tag from all others. At run
time, all class nodes process the given sentence,
applying many classifiers simultaneously. The classifiers then
compete for deciding the pos of this word, and the node that records
the highest activity for a given word in a sentence determines its
pos.
The methods compared use context and collocation features as in \cite{Brill95}.

Given a sentence, each word in the sentence is assigned an initial
tag, based on the most common part of speech in the training corpus.
Then, for each word in the sentence, the network processes the
sentence, and makes a suggestion for the pos of this word. Thus, the
input for the predictor is noisy, since the initial assignment is not
accurate for many of the words.  This process can repeat a few times,
where after predicting the pos of a word in the sentence we re-compute
the new feature-based representation of the sentence and predict
again. Each time the input to the predictors is expected to be
slightly less noisy.  In the results presented here, however, we
present the performance without the re-cycling process, so that we
maintain the linear function expressivity (see \cite{RothZe98} for details).

The results presented in Table~\ref{tab:pos} are based on experiments
using $800,000$ words of the Penn Treebank Tagged WSJ corpus. About
$550,000$ words were used for training and $250,000$ for testing. \snow\
and \tbl\ were trained and tested on the same data.

\begin{table}
\caption{{\bf \pos~ System comparison}.
The first column gives the number of test cases.
All algorithms were trained on $550,000$ words of the tagged WSJ corpus.
Baseline simply predicts according to the most common pos tag for the word
in the training corpus.}
\begin{center}
\begin{tabular}{||l c c c   ||}
\hline
Test    & Baseline &  \tbl & \snow  \\
cases   &          &       &       \\
\hline
250,000 & 94.4     & 96.9  & 96.8  \\
\hline
\end{tabular}
\vspace{-0.3in}
\end{center}
\label{tab:pos}
\end{table}

\section{Conclusion}

We presented an analysis of a few of the commonly used statistics
based and machine learning algorithms for ambiguity resolution tasks.
We showed that all the algorithms investigated can be re-cast as
learning linear separators in the feature space.  We analyzed the
complexity of the function space in which each of these method
searches, and show that they all search a space that is as complex as
the space of all linear separators. We used these to argue motivate 
 our approach of learning a sparse network of linear
separators (\snow), which learns a network of linear separator by
utilizing the Winnow learning algorithm.  
We then
presented an extensive experimental study comparing the \snow\ based
algorithms to other methods studied in the literature on several well
studied disambiguation tasks.  We present experimental results on
\spell, \ppa\ and \pos. In all cases we show that our approach either
outperformed other methods tried for these tasks or performs
comparably to the best.  We view this as a strong evidence to that
this approach provides a unified framework for the study of natural
language disambiguation tasks.

The importance of providing a unified framework stems from the fact
the essentially all ambiguity resolution problems that are addressed
here are at the lower level of the natural language inferences chain.
A large number of different kinds of ambiguities are to be resolved
simultaneously in performing any higher level natural language
inference \cite{Cardie96}.  Naturally, these processes, acting on the
same input and using the same ``memory'', will interact.  A unified
view  of ambiguity resolution
within a single architecture, is valuable if one wants
understand how to put together a large number of these inferences, 
study  interactions among them and make progress towards using these
in performing higher level inferences.

\end{document}